\documentclass[conference]{IEEEtran}
\IEEEoverridecommandlockouts

\usepackage{cite}
\usepackage{amsmath,amssymb,amsfonts}
\usepackage{graphicx}
\usepackage{textcomp}
\def\BibTeX{{\rm B\kern-.05em{\sc i\kern-.025em b}\kern-.08em
    T\kern-.1667em\lower.7ex\hbox{E}\kern-.125emX}}

\usepackage{boldline}
\usepackage{subfig}
\usepackage{bm}

\DeclareMathOperator*{\argmin}{arg\,min} 

\usepackage[noend]{algpseudocode}
\usepackage{algorithm}
\usepackage[hidelinks]{hyperref}
\usepackage[dvipsnames]{xcolor}

\algnewcommand{\LineComment}[1]{\State \(\triangleright\) #1}

\begin{document}

\title{Federated Learning with Diffusion Models \\ for Privacy-Sensitive Vision Tasks


    \thanks{This work was partly supported by the National Research Foundation of Korea(NRF) grant funded by the Korea government(MSIT) (No. RS-2023-00207816), the Institute of Information and Communications Technology Planning and Evaluation (IITP) Grant funded by the Korea Government (MSIT) (Artificial Intelligence Innovation Hub) under Grant 2021-0-02068, (No.RS-2022-00155911, Artificial Intelligence Convergence Innovation Human Resources Development (Kyung Hee University)), (No.2019-0-01287, Evolvable Deep Learning Model Generation Platform for Edge Computing)  and partly by the Korea Institute of Energy Technology Evaluation and Planning(KETEP) and the Ministry of Trade, Industry \& Energy(MOTIE) of the Republic of Korea (No. 20212020800110). *Dr. CS Hong is the corresponding author.}
}


\author{\IEEEauthorblockN{Ye Lin Tun, Chu Myaet Thwal, Ji Su Yoon, Sun Moo Kang, Chaoning Zhang, Choong Seon Hong*}
\IEEEauthorblockA{\textit{Department of Computer Science and Engineering} \\
\textit{Kyung Hee University}\\
Yongin-si, Republic of Korea \\
yelintun@khu.ac.kr, chumyaet@khu.ac.kr, yjs9512@khu.ac.kr, \\etxkang@khu.ac.kr, chaoningzhang1990@gmail.com, cshong@khu.ac.kr}
}



\maketitle

\begin{abstract}

Diffusion models have shown great potential for vision-related tasks, particularly for image generation.
However, their training is typically conducted in a centralized manner, relying on data collected from publicly available sources. 
This approach may not be feasible or practical in many domains, such as the medical field, which involves privacy concerns over data collection.
Despite the challenges associated with privacy-sensitive data, such domains could still benefit from valuable vision services provided by diffusion models.
Federated learning (FL) plays a crucial role in enabling decentralized model training without compromising data privacy.
Instead of collecting data, an FL system gathers model parameters, effectively safeguarding the private data of different parties involved. 
This makes FL systems vital for managing decentralized learning tasks, especially in scenarios where privacy-sensitive data is distributed across a network of clients.
Nonetheless, FL presents its own set of challenges due to its distributed nature and privacy-preserving properties.
Therefore, in this study, we explore the FL strategy to train diffusion models, paving the way for the development of federated diffusion models. 
We conduct experiments on various FL scenarios, and our findings demonstrate that federated diffusion models have great potential to deliver vision services to privacy-sensitive domains. 

\end{abstract}

\vspace{2mm}

\section{Introduction}

Diffusion models~\cite{ho2020denoising} have recently emerged as a powerful tool for image synthesis, offering superior performance compared to previous generative models. Beyond image synthesis, these models can also be applied to vision tasks such as inpainting, colorization, and super-resolution, making them a versatile tool for image manipulation \cite{rombach2022high, saharia2022palette}. The training process of a diffusion model involves adding noise to the image and the model learning to iteratively remove the noise, thereby recovering the original image. Once trained, diffusion models can generate new images from pure noise images by iteratively removing the noise.



Most of the existing studies on diffusion models have been conducted in centralized settings \cite{ho2020denoising, nichol2021improved, song2020denoising, nguyen2023new}. However, collecting training data from multiple sources for centralized training can be expensive and raise privacy concerns. While collecting natural images from public sources can be less invasive, it can raise significant privacy concerns for specific tasks such as medical image synthesis. For instance, even though medical institutions could benefit from cooperative training of medical image diffusion models, exposing private medical records to one another can be a breach of privacy. Therefore, it is essential to explore privacy-preserving training strategies, such as federated learning (FL) \cite{mcmahan2017communication}, for training diffusion models. Such an approach would overcome the privacy and security challenges of centralized training systems, enabling the training of diffusion models over sensitive data.

Federated learning~\cite{mcmahan2017communication} is a distributed training strategy which allows remote client devices to collaboratively train a model. 
In FL, clients train local models on their private data and share the trained model parameters with a central server. 
The server aggregates the received local parameters into a global model that inherits the knowledge learned by the local models. 
In such a way, FL facilitates learning over a distributed network of clients while also preserving their privacy.
Additionally, FL provides significant advantages for network management, particularly in terms of reducing communication bandwidth requirements~\cite{9445589}. This is particularly beneficial in domains such as medical or engineering, where the transmission of training data can be substantial. Furthermore, FL enhances the robustness of distributed computing by leveraging a network of computing devices, reducing dependency on a single centralized computing infrastructure.


FL is often used to train models on the private data of personal devices, such as smartphones.
However, due to the computational expense, training generative diffusion models in the FL context is only feasible for institutional data silos with adequate computing power. 
With the increasing computing power of mobile devices and the development of lightweight diffusion models \cite{10096710}, on-device training of diffusion models using mobile devices may become a possibility in the future.
Training diffusion models using the data at the edge may also improve the quality of personalized services provided to users.

FL comes with its own challenges, particularly with regards to data heterogeneity \cite{mcmahan2017communication}. 
Data residing on client devices (such as data silos or mobile devices) follows different distributions depending on the task and nature of device's usage. 
Training local models on such non-IID data can result in diverse parameters, which can negatively impact the global model's performance when aggregated. 
Moreover, client devices in an FL system can be unreliable, with clients not always available to participate in every training round.  
Therefore, our study aims to investigate how these challenges affect the performance of diffusion models when trained with FL strategies.










\begin{figure*}[htbp]
    \centering
    \subfloat[Federated Learning]{\includegraphics[width=0.28\textwidth]{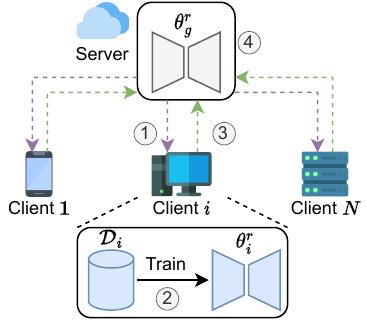} \label{subfig:overview}}
    \subfloat[Local training]{\includegraphics[width=0.58\textwidth]{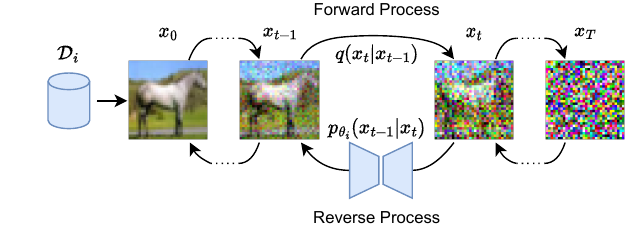} \label{subfig:local_training}} 
	\caption{Training a diffusion model with the federated learning strategy. (a) Overview of a federated learning system. (b) Local training process of a client.}
	\label{fig:system_model}
\end{figure*}

\section{Method}

We employ the widely known FedAvg~\cite{mcmahan2017communication} algorithm for federated learning.
A typical FL process involves a central server and a set of client devices. 
The client devices possess the private data for local model training, while the central server is responsible for coordinating the overall FL process.
The goal of FL is to train a global model $\theta_g$ by optimizing:
\begin{equation}
    \theta_g^* = \argmin_{\theta_g} \sum_{i=1}^N \frac{|\mathcal{D}_i|}{\mathcal{|D|}} \mathcal{L} (\theta_g, \mathcal{D}_i),
    \label{eqn:fl_obj}
\end{equation}
where $\mathcal{L}$ denotes the local loss and $\mathcal{D}_i$ is the dataset of the $i$-th client. Here, $|\mathcal{D}|=\sum_{i=1}^N |\mathcal{D}_i|$ and $N$ is the total number of clients.
As illustrated in Fig.~\ref{subfig:overview}, there are four main steps in each FL communication round $r$: (1) the server sends the global model $\theta^r_g$ to all participating clients, (2) each client $i$ synchronizes the local $\theta_i^r$ with the received global model and trains $\theta_i^r$ on local private data $\mathcal{D}_i$, (3) the trained local parameters are then sent back to the server, (4) the server aggregates the local parameters using weighted averaging to update the global model $\theta^{r+1}_g$. These steps are repeated for a total of $R$ communication rounds.




The local training process for each client $i$ involves forward and reverse diffusion processes \cite{ho2020denoising}, as shown in Fig.~\ref{subfig:local_training}. Given a sample $x_0 \in \mathcal{D}_i$ and the number of timesteps $T$, the forward process $q$ adds a small amount of Gaussian noise to $x_t$ at each timestep $t$ using a pre-defined variance schedule $\beta_t \in (0,1)$. With $\alpha_t := 1-\beta_t$ and $\bar{\alpha}_t := \prod^t_{s=0} \alpha_s$, $x_t$ can be expressed as $x_t = \sqrt{\bar{\alpha}_t} x_0 + \sqrt{1-\bar{\alpha}_t}\epsilon$, where $\epsilon \sim \mathcal{N}(0, I)$.
Given that $T$ is sufficiently large, $x_T$ becomes pure noise. In the reverse process $p_{\theta_i}$, the model predicts the added noise $\epsilon$, which can be removed at each timestep $t$ to recover the original image. The loss function can be expressed as:
\begin{equation}
    L = E_{t, x_0, \epsilon}[||\epsilon - \epsilon_{\theta_i}(x_t, t)||^2].
    \label{eqn:loss_objective}
\end{equation}

After the FL training, the global model $\theta_g$ can iteratively sample $x_{t-1}$ using \eqref{eqn:sample} to obtain a new sample $x_0$, given ${x_T \sim \mathcal{N}(0, I)}$ \cite{ho2020denoising}.
\begin{equation}
    x_{t-1}=\frac{1}{\sqrt{\alpha_t}}(x_t - \frac{1-\alpha_t}{\sqrt{1-\bar{\alpha}_t}} \epsilon_{\theta_g}(x_t, t)) + \sigma_t z.
    \label{eqn:sample}
\end{equation}
Here, $z \sim \mathcal{N}(0, I)$ and  either $\sigma^2_t=\beta_t$ or $\sigma^2_t=\tilde{\beta}_t= \frac{1-\bar{\alpha}_{t-1}}{1-\bar{\alpha}_t}$ can be used \cite{ho2020denoising}. Algorithm~\ref{alg:feddm_training} describes the detailed procedure.






\begin{algorithm}[tbp]
\caption{FedDM}
\label{alg:feddm_training}
\begin{algorithmic}[1]

    \State \textbf{Input:}  number of communication rounds $R$, number of local epochs $E$, number of clients $N$, number of timesteps $T$, learning rate $\eta$, variance schedule $\beta_{1:T}$, ${\alpha_{1:T} := 1-\beta_{1:T}}$
    
    \State \textbf{Output:} global model $\theta^R_{g}$
    \vspace{0.5em}
    
    \State \textbf{Server executes:}
    \State initialize $\theta^0_{g}$
    \For{each round $r=0,1,\dots,R-1$}
        \For{each client $i=1,2,\dots,N$ in parallel}
            \State $\theta^r_i \leftarrow \text{LocalUpdate}(i, \theta^r_g)$
        \EndFor
        \State $\theta^{r+1}_g \leftarrow \sum^N_{i=1} \frac{|\mathcal{D}_i|}{|\mathcal{D}|} \theta^r_i$ 
    \EndFor
    \State return $\theta^R_g$
    \vspace{0.5em}

    \State \textbf{Client executes:} LocalUpdate($i, \theta^r_g$):
    \State $\theta^{r}_i \leftarrow \theta^r_g$
    \For{each epoch $e=0,1,\dots,E-1$}
        \For{each batch $\mathcal{B} \in \mathcal{D}_i$}
            \For{each $x_0 \in \mathcal{B}$ in parallel}
                \State $t \sim \text{Uniform}(\{1, 2, \dots, T\})$
                
                \State $\bar{\alpha}_t := \prod^t_{s=0} \alpha_s$

                \State $\epsilon \sim \mathcal{N}(0, I)$
    

                \State $x_t = \sqrt{\bar{\alpha}_t} x_0 + \sqrt{1-\bar{\alpha}_t}\epsilon$
    
                \State $\mathcal{\ell} \leftarrow ||\epsilon - \epsilon_{\theta_i^r}(x_t, t)||^2$

            \EndFor
            \State $\mathcal{L} = \frac{1}{|\mathcal{B}|} \sum\limits_{x_0 \in \mathcal{B}} \ell$
            \State $\theta^r_i \leftarrow \theta^r_i - \eta \nabla \mathcal{L}$
        \EndFor
    \EndFor
    \State return $\theta^{r}_i$

\end{algorithmic}
\end{algorithm}


    
    

\section{Experiments}

\subsection{Experimental Settings}



In our experiments, we use the following common settings, unless otherwise stated. 
We use the PyTorch~\cite{paszke2019pytorch} framework and follow the diffusion implementation from \cite{Crowson}. We conduct our experiments on the CIFAR-10~\cite{krizhevsky2009learning}, Fashion-MNIST~\cite{xiao2017/online}, and SVHN~\cite{netzer2011reading} datasets, each containing 10 classes.
To simulate the FL scenario, we partition the training portion of each dataset into local data of $10$ clients. To partition the data, we use the Dirichlet distribution~\cite{ferguson1973bayesian}, in which we can adjust the concentration parameter $\beta$ to control the level of clients' data heterogeneity. 
A lower value of $\beta$ indicates a greater degree of data heterogeneity. We set $\beta=0.5$ as the default value.

For our model, we use the U-Net~\cite{ronneberger2015u} backbone containing residual blocks~\cite{he2016deep}. 
We set the number of communication rounds to $300$. 
Each client performs local training for $5$ epochs, using the  Adam optimizer with a learning rate of $2e-4$. 
The input image dimension is set to $32 \times 32$, with a batch size of $512$, and diffusion timesteps of $300$ for all datasets. 
For the CIFAR-10 dataset, we apply a simple random horizontal flip as data augmentation.

To evaluate the quality of the generated samples, we use the Inception Score (IS)~\cite{salimans2016improved} and Fréchet Inception Distance (FID)~\cite{heusel2017gans}. For evaluation, we generate the same number of samples as in the testing set of each dataset. Our FID scores are computed with respect to both the training and testing sets.



\subsection{Comparison with Centralized Training}
\label{sec:centralized_training}

For the centralized training approach, we use the original training portion of each dataset. We set the number of epochs to $1500$ to ensure that the centralized model receives the same amount of training as the FL model. Specifically, we calculate the number of epochs for centralized training as $E_{\text{cen}} = R \times E$, where $R$ is the total number of FL communication rounds and $E$ is the number of local epochs.

\begin{figure*}[tb]
    \centering
    \subfloat[Centralized]{\includegraphics[width=0.14\textwidth]{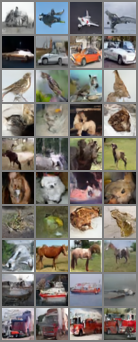} \label{subfig:cen_cifar10}} 
    \subfloat[FedAvg]{\includegraphics[width=0.14\textwidth]{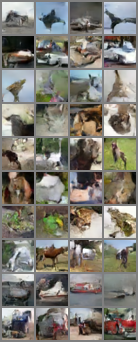} \label{subfig:fl_cifar10}} 
    \hspace{1em}
    \subfloat[Centralized]{\includegraphics[width=0.14\textwidth]{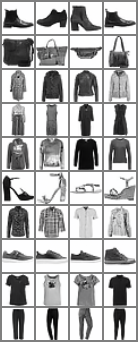} \label{subfig:cen_fashionmnist}}
    \subfloat[FedAvg]{\includegraphics[width=0.14\textwidth]{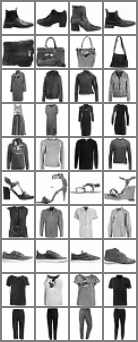} \label{subfig:fl_fashionmnist}}
    \hspace{1em}
    \subfloat[Centralized]{\includegraphics[width=0.14\textwidth]{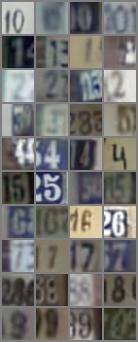} \label{subfig:cen_svhn}} 
    \subfloat[FedAvg]{\includegraphics[width=0.14\textwidth]{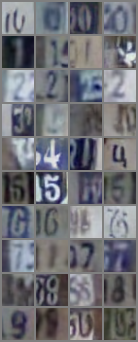} \label{subfig:fl_svhn}}
    \caption{Generated samples for CIFAR-10, Fashion-MNIST and SVHN datasets using centralized training and federated learning strategies. Each row represents a different class.}
    \label{fig:gen_cen_fl}
\end{figure*}

Figure~\ref{fig:gen_cen_fl} compares the quality of samples generated by the centralized and FL models across different datasets, while Table~\ref{tab:is_fid_diff_dataset} reports their respective IS and FID metrics. 
The results in Table~\ref{tab:is_fid_diff_dataset} indicate a significant decline in the quality of samples generated by the FL global model compared to the centralized model.
However, upon visual examination of Fig.~\ref{fig:gen_cen_fl}, it is observed that while samples from the centralized model may appear slightly more visually appealing for the CIFAR-10 dataset, those from the FL global model for the Fashion-MNIST and SVHN datasets can be relatively comparable, albeit with some degree of blurriness. These findings highlight the potential of training diffusion models in FL settings and the need for further exploration of improved FL designs for such training.

\begin{table*}[tb]
\centering
\caption{IS and FID for centralized and FL approaches.}
\label{tab:is_fid_diff_dataset}
\scalebox{0.95}{
\begin{tabular}{lV{3}cccV{3}cccV{3}ccc}
\hlineB{3}
                     & \multicolumn{3}{cV{3}}{\textbf{IS $\uparrow$}}                        & \multicolumn{3}{cV{3}}{\textbf{FID (train) $\downarrow$}} & \multicolumn{3}{c}{\textbf{FID (test) $\downarrow$}} \\
                     & CIFAR-10             & Fashion-MNIST        & SVHN                 & CIFAR-10          & Fashion-MNIST   & SVHN             & CIFAR-10         & Fashion-MNIST   & SVHN            \\ \hline
\textbf{Centralized} & \textbf{4.45 ± 0.12} & \textbf{4.15 ± 0.08} & \textbf{3.39 ± 0.04} & \textbf{10.30}    & \textbf{7.93}   & \textbf{13.18}   & \textbf{11.70}   & \textbf{8.92}   & \textbf{13.13}  \\
\textbf{FedAvg}      & 3.69 ± 0.08          & 4.11 ± 0.12          & 3.07 ± 0.04          & 21.42             & 17.03           & 15.64            & 22.67            & 17.91           & 20.39           \\ \hlineB{3}
\end{tabular}
}
\end{table*}


\begin{figure*}[tb]
    \centering
    \subfloat[$\beta=0.1$]{\includegraphics[width=0.26\textwidth]{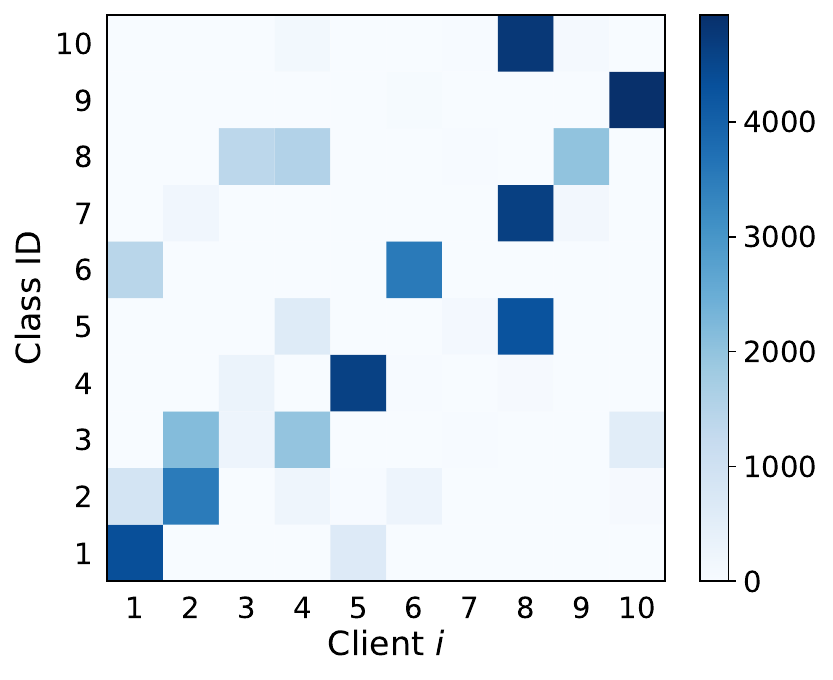} \label{subfig:beta_0.1}}
	\subfloat[$\beta=0.5$]{\includegraphics[width=0.26\textwidth]{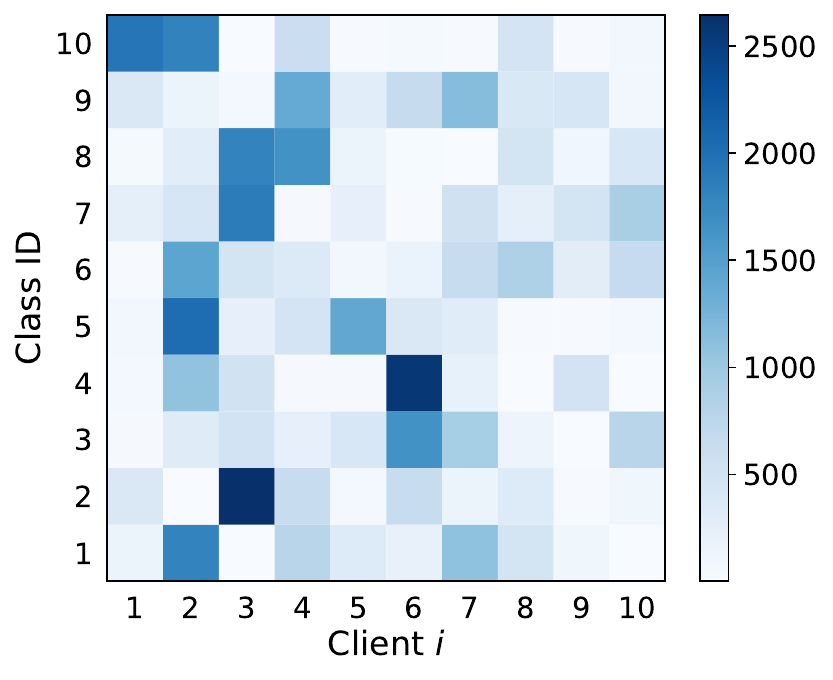} \label{subfig:beta_0.5}} 
    \subfloat[$\beta=5.0$]{\includegraphics[width=0.26\textwidth]{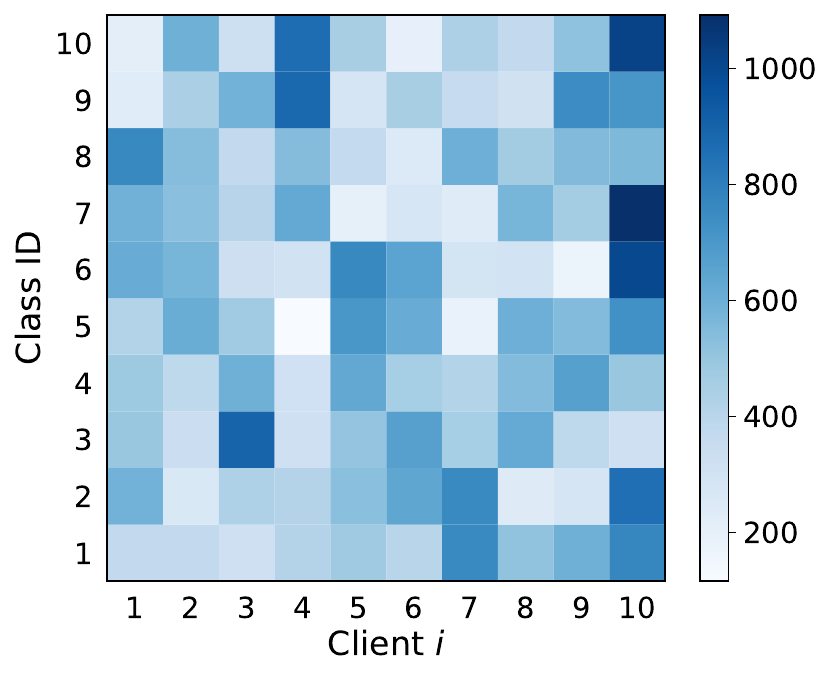} \label{subfig:beta_5}} 
	\caption{Client data distribution with different $\beta$ values for the CIFAR-10 dataset. A darker rectangle denotes a higher number of data samples for a specific class in a client.}
	\label{fig:betas_distribution}
\end{figure*}

\subsection{Data Heterogeneity}

To investigate the effects of data heterogeneity on the quality of generated samples, we perform FL training with $\beta$ values of $0.1$, $0.5$, and $5$. 
Figure~\ref{fig:betas_distribution} illustrates the clients' data distributions for each $\beta$ value, while Table~\ref{tab:is_fid_beta} presents the IS and FID metrics of the generated samples.
Although the IS values are relatively comparable across the settings, the FID scores are more sensitive to data heterogeneity. 
For each dataset, $\beta=0.1$ with the highest degree of data heterogeneity achieves the worst FID score, while $\beta=5$ achieves the best.
Overall, the quality of samples decreases as the degree of data heterogeneity increases, since local training tends to drift towards the local data distribution. 
This highlights the need for further investigation into whether existing FL techniques for combating data heterogeneity can be applied to training diffusion models.

\begin{table*}[tbp]
\centering
\caption{IS and FID for different $\beta$ values.}
 \label{tab:is_fid_beta}
 \scalebox{0.95}{
\begin{tabular}{lV{3}cccV{3}cccV{3}ccc}
\hlineB{3}
                           & \multicolumn{3}{cV{3}}{\textbf{IS $\uparrow$}}                                   & \multicolumn{3}{cV{3}}{\textbf{FID (train) $\downarrow$}}        & \multicolumn{3}{c}{\textbf{FID (test) $\downarrow$}}          \\
                           & CIFAR-10             & Fashion-MNIST        & SVHN                 & CIFAR-10       & Fashion-MNIST  & SVHN           & CIFAR-10       & Fashion-MNIST  & SVHN           \\ \hlineB{3}
\textbf{\pmb{$\beta=0.1$}} & 3.65 ± 0.11          & 3.94 ± 0.09          & \textbf{3.14 ± 0.05} & 25.40          & 20.01          & 20.15          & 26.74          & 20.93          & 22.93          \\
\textbf{\pmb{$\beta=0.5$}} & \textbf{3.69 ± 0.08} & \textbf{4.11 ± 0.07} & 3.07 ± 0.04          & 21.42          & 17.03          & 15.64          & 22.67          & 17.91          & 20.39          \\
\textbf{\pmb{$\beta=5$}}   & 3.66 ± 0.10          & 3.93 ± 0.08          & 3.12 ± 0.06          & \textbf{20.57} & \textbf{15.58} & \textbf{12.28} & \textbf{21.91} & \textbf{16.58} & \textbf{16.38} \\ \hlineB{3}
\end{tabular}
}
\end{table*}

\subsection{Number of Clients}

We conduct experiments by varying the number of clients, setting $N$ to $10$, $30$, and $50$. It is important to note that the total amount of data samples is the same across all settings, and increasing the number of clients results in a more distributed data setup. While all clients participate in each communication round for $N=10$, we randomly sample 10 clients to participate for $N=30$ and $N=50$ settings, which reflects the practical FL scenarios where not all clients are available to participate in every round. As shown in Table~\ref{tab:is_fid_num_clients}, the quality of the generated samples decreases as the number of clients increases, with $N=50$ having the worst FID scores compared to the other settings. In our experiment, we mainly aim to investigate the effects of data distributed over a large number of clients. 
However, it is worth mentioning that in practice, more clients could lead to an improved quality of the FL global model due to the increased number of training data samples.





\begin{table*}[tbp]
\centering
\caption{IS and FID for different number of clients.}
\label{tab:is_fid_num_clients}
\scalebox{0.95}{
\begin{tabular}{lV{3}cccV{3}cccV{3}ccc}
\hlineB{3}
                      & \multicolumn{3}{cV{3}}{\textbf{IS $\uparrow$}}                        & \multicolumn{3}{cV{3}}{\textbf{FID (train) $\downarrow$}} & \multicolumn{3}{c}{\textbf{FID (test) $\downarrow$}} \\
                      & CIFAR-10             & Fashion-MNIST        & SVHN                 & CIFAR-10         & Fashion-MNIST    & SVHN             & CIFAR-10         & Fashion-MNIST   & SVHN            \\ \hlineB{3}
\textbf{\pmb{$N=10$}} & \textbf{3.69 ± 0.08} & 4.11 ± 0.07          & \textbf{3.07 ± 0.04} & \textbf{21.42}   & \textbf{17.03}   & \textbf{15.64}   & \textbf{22.67}   & \textbf{17.91}  & \textbf{20.39}  \\
\textbf{\pmb{$N=30$}} & 3.02 ± 0.05          & 3.98 ± 0.11          & 2.95 ± 0.04          & 40.09            & 19.56            & 20.45            & 41.57            & 20.50           & 23.48           \\
\textbf{\pmb{$N=50$}} & 2.81 ± 0.07          & \textbf{4.21 ± 0.07} & 2.96 ± 0.04          & 48.69            & 35.61            & 23.76            & 50.23            & 36.46           & 26.46           \\ \hlineB{3}
\end{tabular}
}
\end{table*}

\subsection{Number of Local Epochs}

We also investigate the impact of the number of local epochs on model performance by setting $E$ to $1$, $5$, and $10$. The results are shown in Table~\ref{tab:is_fid_local_epochs}, where we observe that the quality of generated samples generally improves with an increase in the number of local epochs. This  suggests that longer training is beneficial for diffusion models with a large number of parameters. However, client devices in an FL environment typically have limited computing and communication resources. Therefore, increasing the number of local epochs can result in more computing burden on the clients. Moreover, this could also increase the risk of clients dropping out of the FL process without completing local training. Alternatively, we could decrease the number of local epochs while increasing the total number of communication rounds to achieve similar results. However, this approach would require significant communication resources for both the server and clients. Therefore, developing lightweight diffusion models that can facilitate training and deployment on edge devices is an important topic for future studies.

\begin{table*}[tb]
\centering
\caption{IS and FID scores for different local epochs.}
\label{tab:is_fid_local_epochs}
\scalebox{0.95}{
\begin{tabular}{lV{3}cccV{3}cccV{3}ccc}
\hlineB{3}
                     & \multicolumn{3}{cV{3}}{\textbf{IS $\uparrow$}}                        & \multicolumn{3}{cV{3}}{\textbf{FID (train) $\downarrow$}} & \multicolumn{3}{c}{\textbf{FID (test) $\downarrow$}} \\
                     & CIFAR-10             & Fashion-MNIST        & SVHN                 & CIFAR-10         & Fashion-MNIST    & SVHN             & CIFAR-10         & Fashion-MNIST   & SVHN            \\ \hlineB{3}
\textbf{\pmb{$E=1$}}  & 2.49 ± 0.04          & 4.01 ± 0.12          & 2.33 ± 0.03          & 79.79            & 36.36            & 48.16            & 81.60            & 37.31           & 52.85           \\
\textbf{\pmb{$E=5$}}  & 3.69 ± 0.08          & \textbf{4.11 ± 0.07} & 3.07 ± 0.04          & 21.42            & 17.03            & 15.64            & 22.67            & 17.91           & 20.39           \\
\textbf{\pmb{$E=10$}} & \textbf{3.88 ± 0.12} & 3.95 ± 0.07          & \textbf{3.10 ± 0.04} & \textbf{18.22}   & \textbf{12.78}   & \textbf{11.73}   & \textbf{19.51}   & \textbf{13.75}  & \textbf{17.31}  \\ \hlineB{3}
\end{tabular}
}
\end{table*}

\begin{figure*}[htbp]
    \centering
    \subfloat[Client 1]{\includegraphics[width=0.28\textwidth]{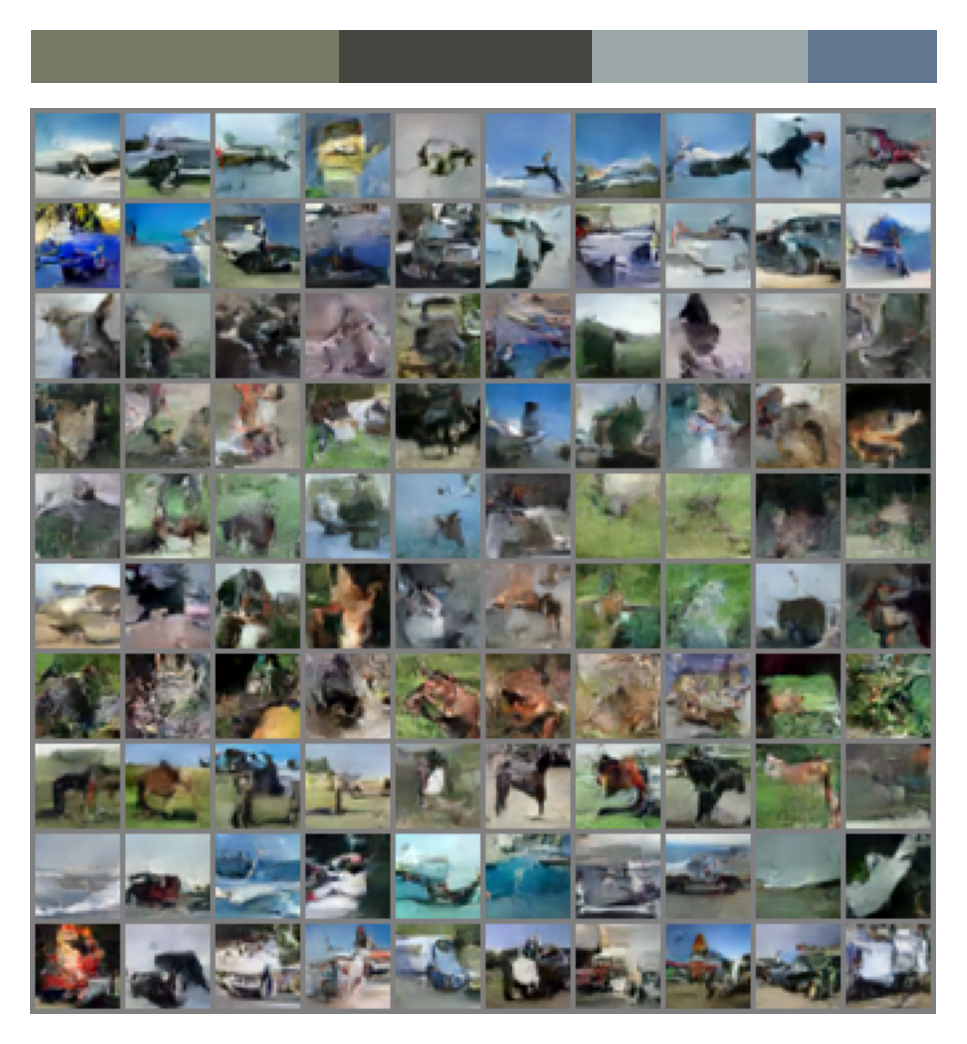} \label{subfig:cifar_c1_color}} 
	\subfloat[Client 2]{\includegraphics[width=0.28\textwidth]{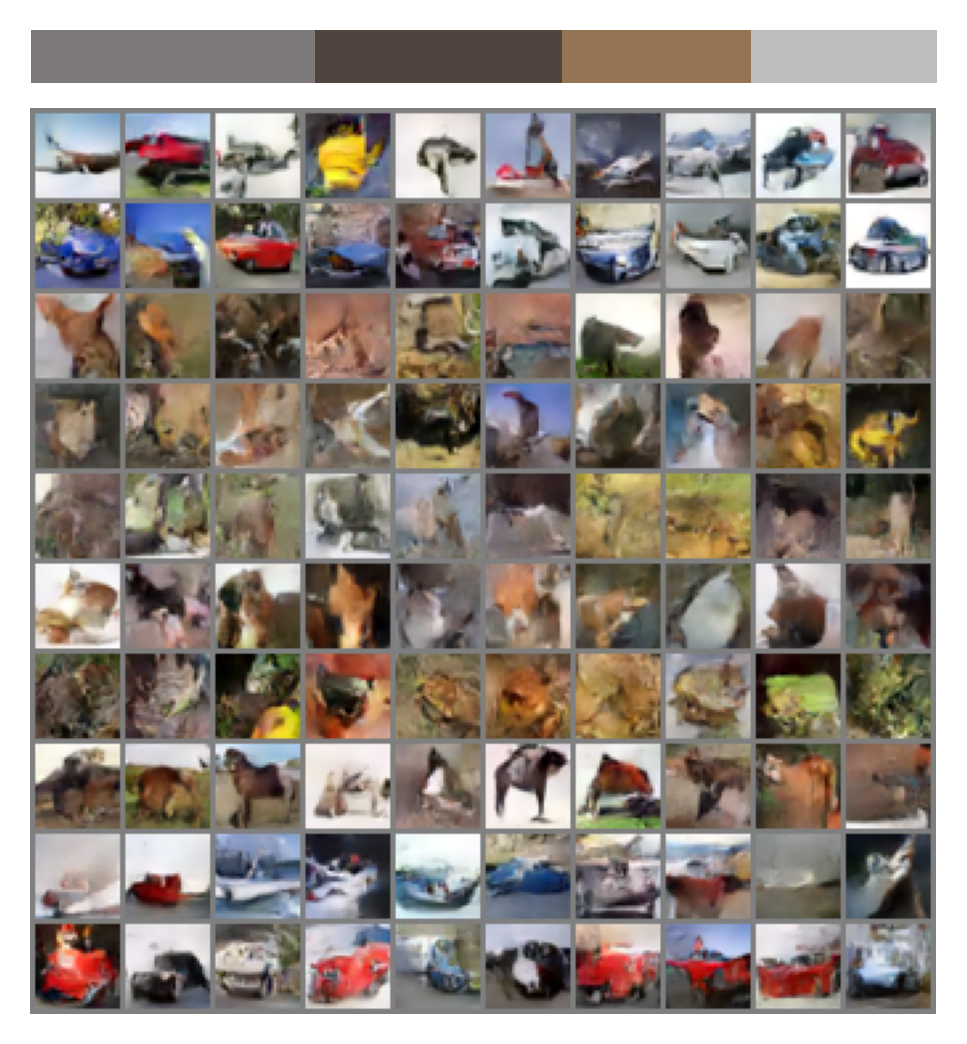} \label{subfig:cifar_c2_color}} 
	\subfloat[Client 5]{\includegraphics[width=0.28\textwidth]{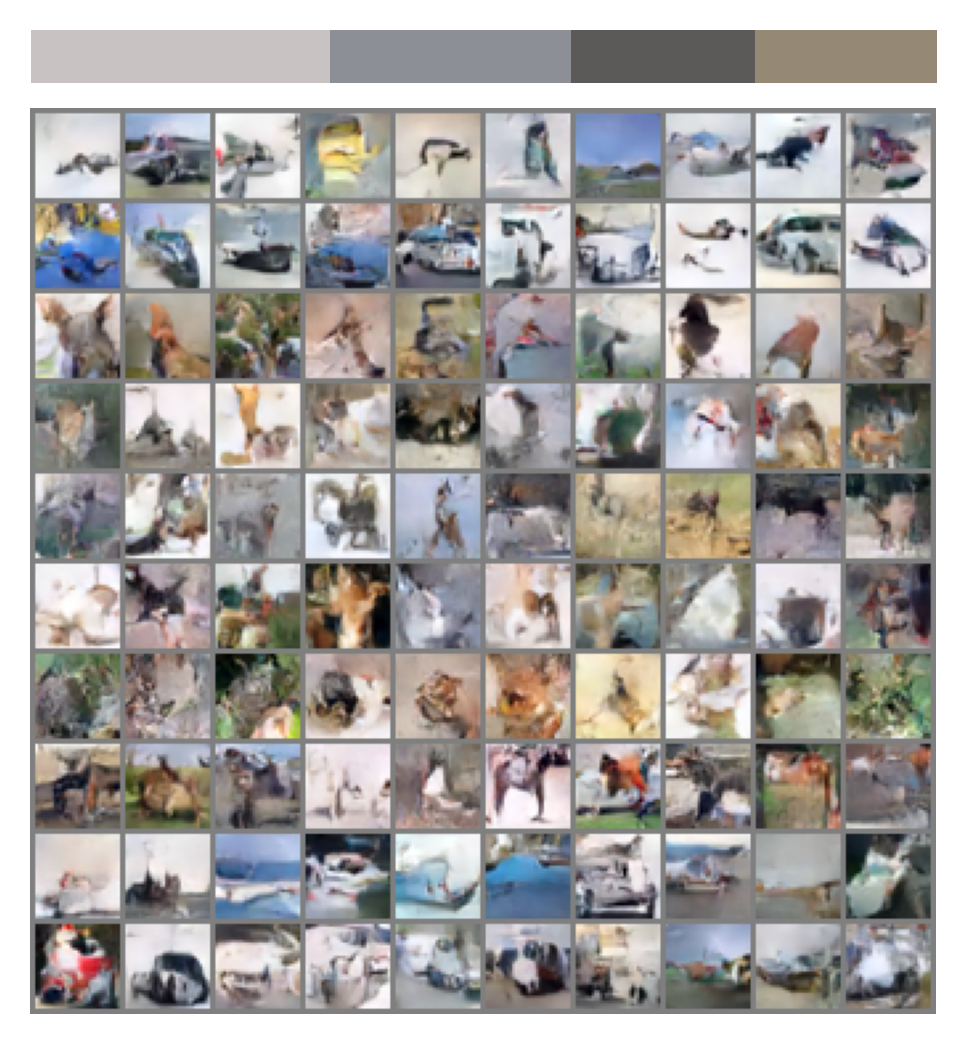} \label{subfig:cifar_c5_color}} 
	\caption{CIFAR-10 samples generated by local models of different clients. Each row represents a different class.}
	\label{fig:gen_client_cifar}
\end{figure*}

\begin{figure*}[htbp]
    \centering
    \subfloat[Client 3]{\includegraphics[width=0.35\textwidth]{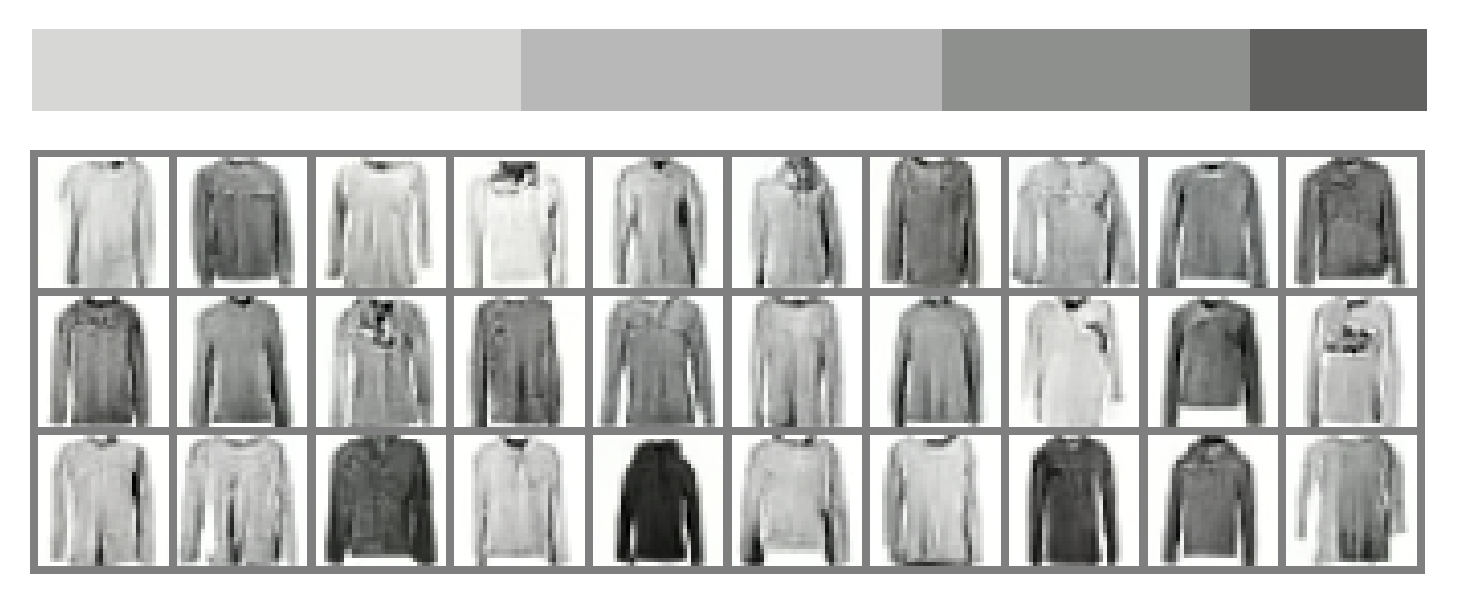} \label{subfig:fashion_c3_color}}
    \hspace{1em}
	\subfloat[Client 6]{\includegraphics[width=0.35\textwidth]{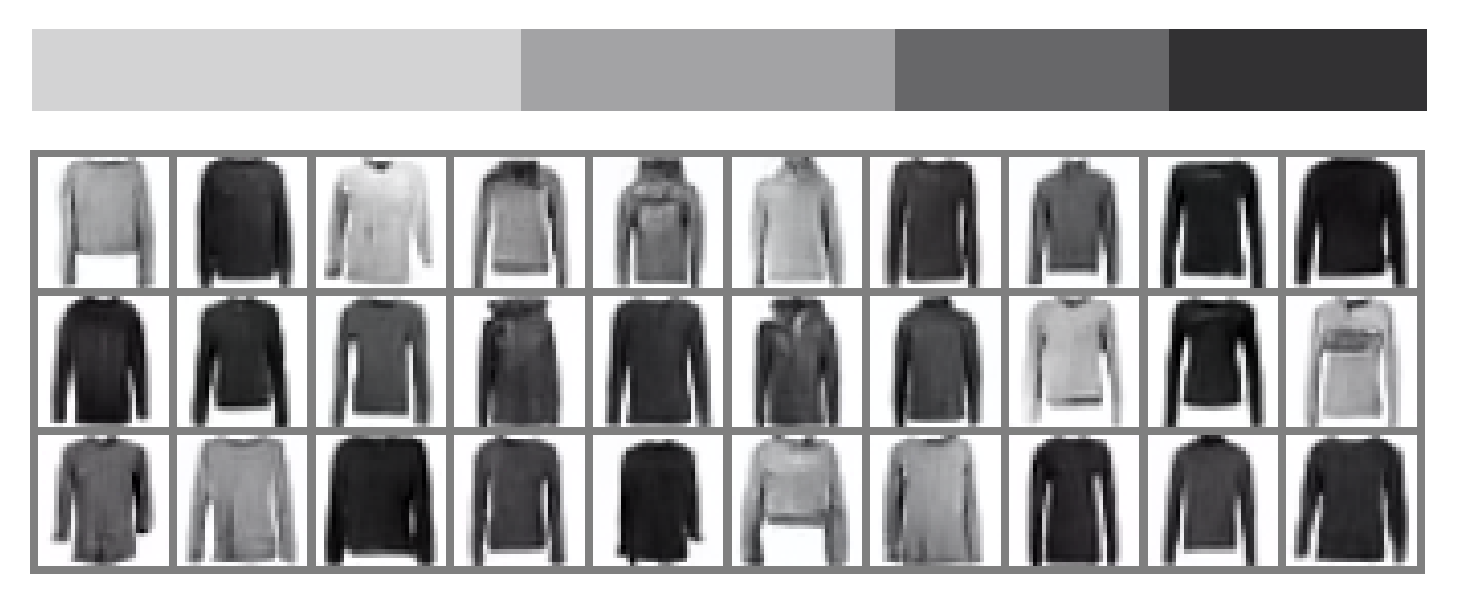} \label{subfig:fashion_c6_color}} 
	\caption{Fashion-MNIST samples of a random class generated by local models of different clients.}
	\label{fig:gen_client_fashion}
\end{figure*}

\subsection{Personalized Local Models}

In an FL environment, clients may have distinct data distributions, leading to local models being biased towards local data. While this can challenge model aggregation and affect global model performance in early FL rounds, these local models can act as personalized models in later rounds. Therefore, we investigate whether these local models have such personalized properties by comparing their generated samples at round $r=300$. Figures~\ref{fig:gen_client_cifar} and \ref{fig:gen_client_fashion} present the samples generated by local models for CIFAR-10 and Fashion-MNIST datasets, with a summary of their color clusters on top. Observing Fig.~\ref{subfig:cifar_c1_color} and its color clusters reveals that samples from client 1 contain a bluish tone that is relatively absent in other clients. A similar observation can be made for Fig.~\ref{subfig:cifar_c2_color} with a brownish tone. Meanwhile, samples from client 5 in Fig.~\ref{subfig:cifar_c5_color} have a relatively lighter color compared to the samples from other clients. For the Fashion-MNIST dataset, samples from client 3 have a relatively lighter color, while those from client 6 have a darker tone. While these findings suggest that the local models are indeed biased towards their own data distributions, they can also provide an opportunity to offer personalized services to clients.

\begin{table}[htbp]
\centering
\caption{IS and FID scores for the SARS-CoV-2 CT-scan dataset.}
\label{tab:is_fid_sar}
\begin{tabular}{lV{3}cV{3}c}
\hlineB{3}
\textbf{}            & \textbf{IS $\uparrow$} & \textbf{FID $\downarrow$} \\ \hlineB{3}
\textbf{Centralized} & \textbf{1.72 ± 0.04}   & \textbf{48.17}            \\
\textbf{FedAvg}          & 1.60 ± 0.03              & 134.98                    \\ \hlineB{3}
\end{tabular}
\end{table}

\subsection{Medical Image Synthesis}

We conduct an experiment on medical image synthesis using the SARS-CoV-2 CT-scan dataset~\cite{soares2020sars}, which contains a total of $2482$ CT scans. The dataset comprises $1252$ scans from patients infected with SARS-CoV-2 and $1230$ scans from non-infected patients with other pulmonary diseases. We perform the training using both centralized and FL approaches, with default parameters except for the image size set to $64 \times 64$ and the batch size set to $128$.

\begin{figure}[tbp]
    \centering
    \subfloat[Source]{\includegraphics[width=0.4\textwidth]{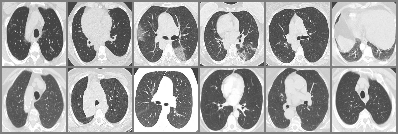} \label{subfig:sar_org}}
    \\
	\subfloat[Centralized]{\includegraphics[width=0.4\textwidth]{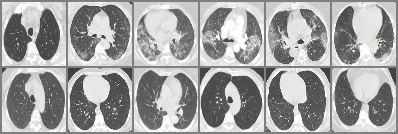} \label{subfig:sar_cen}} 
    \\
    \subfloat[FedAvg]{\includegraphics[width=0.4\textwidth]{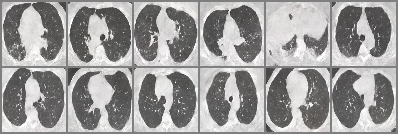} \label{subfig:sar_fl}} 
	\caption{Source and generated samples for the SARS-CoV-2 CT-scan dataset.}
	\label{fig:sar_samples}
\end{figure}

We report the IS and FID metrics in Table~\ref{tab:is_fid_sar}, and Fig.~\ref{fig:sar_samples} compares the generated samples with those from the source dataset. While Table~\ref{tab:is_fid_sar} indicates that the centralized model generates samples with superior quality compared to the FL model, Fig.~\ref{fig:sar_samples} shows that the generated samples from both models can be perceptually similar. This demonstrates a great potential of federated diffusion models for the medical field, where privacy concerns are prominent. However, it is important to mention that the generated samples have not yet been evaluated by a medical expert, which is necessary for further assessment.


\section{Conclusion}

Diffusion models can be effective tools for vision tasks in privacy-sensitive domains, particularly in the medical field. 
However, collecting training data from such domains can be impractical. 
To address this challenge, we explore the federated learning strategy for training diffusion models. 
Our experiments demonstrate that FL strategies hold great potential for managing the training of diffusion models on privacy-sensitive data that is distributed over a network of clients.
As a future direction, it would be interesting to investigate the applicability of existing techniques for mitigating FL challenges, such as data heterogeneity, to federated diffusion models.

\bibliographystyle{IEEEtran}
\bibliography{main}

\end{document}